\documentclass[letterpaper, 10 pt, journal, twoside]{ieeetran}

\usepackage{graphicx}
\usepackage{amsmath}
\usepackage{amssymb}
\usepackage{cite}
\usepackage[letterpaper,top=54pt,bottom=54pt,left=54pt,right=54pt]{geometry}
\usepackage{epstopdf}
\usepackage[lofdepth,lotdepth]{subfig}
\usepackage{booktabs}
\usepackage{color,soul}

\title{
	A Robust Controller for Stable 3D Pinching Using~Tactile~Sensing
	\vspace{0em}
}

\author{Efi Psomopoulou$^{1}$, Nicholas Pestell$^{1}$, Fotios Papadopoulos$^{2}$, John Lloyd$^{1}$, Zoe Doulgeri$^{3}$ and Nathan~F.~Lepora$^{1}$%
\thanks{Manuscript received: February, 24, 2021; Revised April, 27, 2021; Accepted June, 24, 2021.}%
\thanks{This paper was recommended for publication by Editor Dan Popa upon evaluation of the Associate Editor and Reviewers' comments.
This work was supported by a Leverhulme Leadership Award on an “A biomimetic forebrain for robot touch” (RL-2016-39). (\textit{Corresponding author: Efi Psomopoulou.})}
\thanks{$^{1}$EP, NP, JL and NFL are with the Department of Engineering Mathematics and Bristol Robotics Laboratory, University of Bristol, Bristol BS8 1QU, U.K.
        {\tt\footnotesize efi.psomopoulou@bristol.ac.uk; n.pestell@bristol.ac.uk; jl15313@bristol.ac.uk; n.lepora@bristol.ac.uk}}%
\thanks{$^{2} $FP is with the Shadow Robotics Company, London NW5 1LP, U.K.
        {\tt\footnotesize fotios@shadowrobot.com}}%
\thanks{$^{3} $ZD is with the Department of Electrical and Computer Engineering, Aristotle University of Thessaloniki, 54124 Thessaloniki, Greece.
        {\tt\footnotesize doulgeri@ece.auth.gr}}%
\thanks{Digital Object Identifier (DOI): see top of this page.}
}

\begin{document}
	
	\maketitle
	
	\begin{abstract}
		This paper proposes a controller for stable grasping of unknown-shaped objects by two robotic fingers with tactile fingertips. The grasp is stabilised by rolling the fingertips on the contact surface and applying a desired grasping force to reach an equilibrium state. The validation is both in simulation and on a fully-actuated robot hand (the Shadow Modular Grasper) fitted with custom-built optical tactile sensors (based on the BRL TacTip). The controller requires the orientations of the contact surfaces, which are estimated by regressing a deep convolutional neural network over the tactile images. Overall, the grasp system is demonstrated to achieve stable equilibrium poses on various objects ranging in shape and softness, with the system being robust to perturbations and measurement errors. This approach also has promise to extend beyond grasping to stable in-hand object manipulation with multiple fingers. 
		\vspace{-5pt}
	\end{abstract}

	\section{INTRODUCTION}
	\IEEEPARstart{A}{chieving} human-like dexterity with robot hands has been a major goal of robotics for many years. Various novel robot hands have been built, often focusing on their morphology to resemble the human hand and imitate its functionality with use of soft materials and underactuation \cite{doi:10.1146/annurev-control-060117-105003,Shadow,doi:10.1177/0278364915592961,doi:10.1177/0278364913518998}. However, our understanding of how signals interact between the brain, the sensory system and the fingers is still limited. Several review papers summarising progress on multi-finger robot grasping and dexterous object manipulation have been published in the last decade \cite{Bohg2013,Billardeaat8414}. A central message is that it remains a challenge to combine all of the mechanical and tactile sensing components necessary to achieve the performance of the human hand with robots.
	
	Although robotic hands with integrated soft tactile sensors are becoming more common, almost all of the research focuses on object recognition, slip detection and grasp success prediction \cite{KAPPASSOV2015195,YOUSEF2011171,james2020tactile}, with little work on applying tactile sensing to achieve or improve grasp stability. However, grasp stability and manipulation dexterity appear to be closely connected to the rolling ability of human fingertips, which allows for fine adjustment of the contact positions with an object \cite{arimotobook,doi:10.1080/01691864.2017.1365011}. Therefore, being able to robustly control fingertip rolling via proprioceptive and tactile feedback could bridge the gap between human and robot grasping/manipulation in unstructured and unknown environments.
	
	\begin{figure}[t]
		\vspace{-.5em}
        \centering
		\begin{tabular}{@{}c@{}}
			\includegraphics[width=0.85\columnwidth,trim={0 0 0 0},clip]{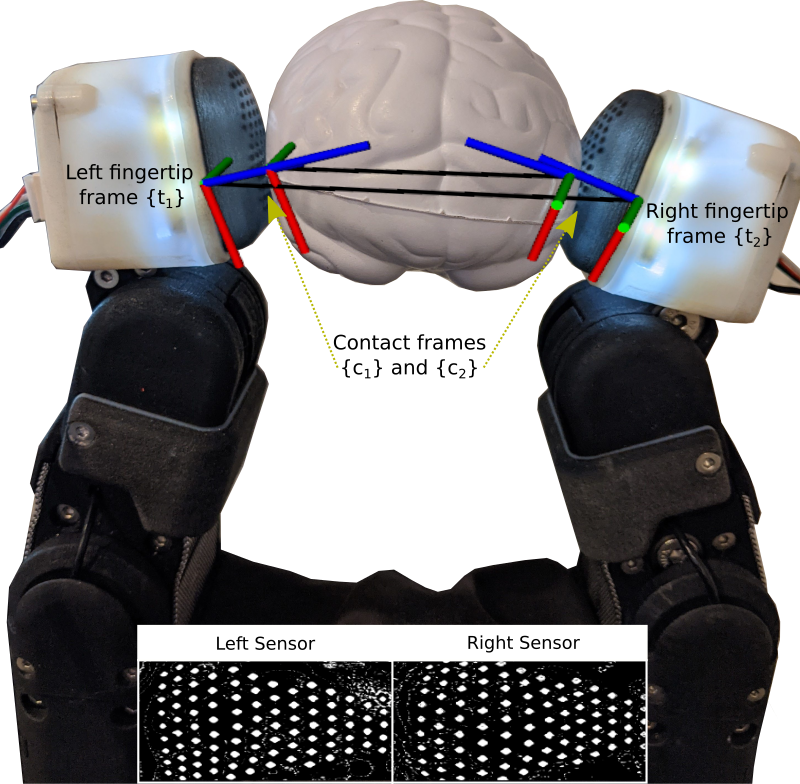}
		\end{tabular}
		\vspace{-5pt}
		\caption{The Shadow Modular Grasper, with two tactile fingertips, stably grasping a brain-shaped stress toy. The controller has rolled the fingertips so that the two lines connecting the fingertips and the contact points are parallel. Tactile images from the two fingertips are also shown.}
		\vspace{-1em}
		\label{fig1}	
	\end{figure}
	
	The present work makes progress in this direction by proposing and implementing a control law that makes use of tactile feedback to roll the fingertips of two robotic fingers on an object's surface into a stable grasp. We focus on the grasp itself, assuming that the robot hand has already reached the object in a ready-to-grasp configuration. Our previous work focused on the  planar case where no tactile sensing was needed \cite{psomopoulou_karashima_doulgeri_tahara_2018}. Here we extend the controller to 3D where tactile information is necessary to achieve grasp stability. The proposed control law allows pinching of an arbitrary-shaped object, using tactile information about the contact orientation to attain a stable equilibrium state by fingertip-rolling motions. No knowledge of the system dynamics is needed nor any trajectory planning performed for the finger movement. A preset desired grasping force is further achieved and the control law can be extended to achieve in-hand manipulation using touch. 
	
	The main contributions of this work concern a robust controller for stable pinch grasping using tactile sensing:\\
	\noindent 1) We implement a low-complexity controller that robustly achieves grasp stability with two fingers, using only proprioceptive and contact information from tactile sensing.\\
	\noindent 2) The controller is validated by both simulations and experiments on novel objects with a fully-actuated hand (Shadow Modular Grasper) with custom tactile-sensing fingertips.\\
	\noindent 3) We show that the system dynamically achieves an equilibrium with a pre-set grasping force from varied initial configurations, assuming it is reachable by a rolling motion. 
	
	\section{BACKGROUND AND RELATED WORK} \label{bgd}
	Early research work on grasp control has focused on detailed grasp analysis and planning of form and force closure grasps involving the accurate planning of fixed contact locations and contact forces \cite{Prattichizzo2008,Wimbock2011,Prattichizzo2012}. More recently, the availability of grasp planning simulators has helped data-driven methods become popular. These approaches rely on sampling grasp candidates from some knowledge base and ranking them according to a specified metric \cite{Goldfeder2007,Ciocarlie2009}. All the above approaches are characterised by static analysis. 
	
	Although force closure implies the existence of an equilibrium, this is not sufficient for ensuring grasp stability \cite{Prattichizzo2008}. The need for further study of grasp dynamics and the development of analytical models that better resemble reality is identified in Bohg \textit{et al.} \cite{Bohg2013}. One way to work towards this is to design model-free grasp controllers that dynamically achieve a stable grasp equilibrium state. Previous research work in this direction includes feedback control laws of low complexity based on rolling contacts \cite{Grammatikopoulou2014,doi:10.1080/01691864.2017.1365011}. 
	
	Grasp control needs state information about the contact, which is provided by tactile sensing at the fingertips. There has been much work on tactile sensing in robot hands using capacitive sensors, tactile skin, BioTac sensors, barometric sensors \cite{6584238,DBLP:journals/corr/abs-1806-05031,doi:10.1177/0278364913514466} or optical tactile sensors, such as the GelSight \cite{yuan_gelsight_2017} and the BRL TacTip \cite{doi:10.1089/soro.2017.0052,lepora2021soft}. This work focuses on the TacTip because it has been integrated recently into a fully-actuated hand, the Shadow Modular Grasper \cite{8656557}. The TacTip is a biomimetic optical tactile sensor based on the dermal papillae structure in human tactile skin, and is fabricated by 3D-printing pinlike structures in a compliant skin imaged with a camera. Estimating contact state information, such as the pose of an object's surface, has been recently demonstrated using deep learning on the tactile images \cite{9058673}.
	
	This paper focuses on controllers that dynamically achieve a stable grasp equilibrium state. It extends previous work that focused on a planar problem of achieving a stable grasp and desired relative finger orientation by rolling fingertips on the object's surface, the direction of which is singular~\cite{psomopoulou_karashima_doulgeri_tahara_2018}. This work considers the full 3D case where system dynamics are more complex and rolling motion on the contact surface has two degrees of freedom. As the complexity of the problem increases, the need for tactile information arises. By using a tactile sensor which accurately perceives the orientation of the contacted surface, such as the TacTip, we propose a robust controller that dynamically achieves grasp stability. 
	
	\section{METHODS}

	\subsection{System modelling} \label{model}

	Consider two robotic fingers of $n_i$ degrees of freedom (DOF), labelling the fingers $i=1,2$, with revolute joints and rigid hemispherical tips of radius $r_1=r_2=r$ in contact with a rigid object of arbitrary shape. This paper makes the following assumptions:\\
	\noindent i) An equilibrium state is assumed reachable by fingertip rolling motion on the object surface.\\
	\noindent ii) Friction at the contacts is sufficient to sustain tangential contact forces so that a rolling constraint holds at all times.\\
    \noindent iii) Both fingertips and the objects are assumed rigid.\\
	\noindent iv) The mass of the object is small enough to ignore gravity.
	
The vector \begin{math}\mathbf{q}_i = \begin{bmatrix} q_{i1} & q_{i2} & \ldots & q_{in_i} \end{bmatrix}^T\end{math} denotes the joint angles for the $i_{th}$ finger and \begin{math} \mathbf{p}_{o}\end{math} $\in\mathbb{R}^{3}$ is the object's position vector. Let \begin{math} \mathbf{p}_{c_i}\end{math} $\in\mathbb{R}^{3}$ describe the position vector of each finger's contact point with the object. The contact frame's orientations are defined by the normal unit vector pointing inwards \begin{math} \mathbf{n}_{z_i}\end{math} $\in\mathbb{R}^{3}$ and the tangential unit vectors to the contact surface, $\mathbf{t}_{x_{i}}, \mathbf{t}_{y_{i}}$  $\in\mathbb{R}^{3}$ (Fig. \ref{f2}).
		
	The system is modelled under the following contact and rolling constraints:
	\begin{align}
		\begin{bmatrix} D_{ii} & D_{i3} \end{bmatrix} \begin{bmatrix} \mathbf{\dot{q}}_i \\ \mathbf{\dot{p}}_o \\ \boldsymbol{\omega}_o \end{bmatrix} = 0 , \, \,
		\begin{bmatrix} A_{ii} & A_{i3} \end{bmatrix} \begin{bmatrix} \mathbf{\dot{q}}_i \\ \mathbf{\dot{p}}_o \\ \boldsymbol{\omega}_o \end{bmatrix}  = 0 ,\label{cons}
	\end{align}
	where $\mathbf{\dot{q}}_{i}\in\mathbb{R}^{n_i}$ are the joint velocities of the $i_{th}$ finger, $\mathbf{\dot{p}}_{o}$, $\boldsymbol{\omega}_o$ are the object's translational and rotational velocities, and
	\begin{align*}
		D_{ii} & = \mathbf{n}_{z_i}^T J_{v_i}, \! \! & D_{i3} =  \begin{bmatrix} -\mathbf{n}_{z_i}^T & \mathbf{n}_{z_i}^T \hat{p}_{{oc}_i}\end{bmatrix}\\
		A_{ii} & = \begin{bmatrix}
			\mathbf{t}_{x_{i}}^T {J_{v_i}} + r \mathbf{t}_{y_{i}}^T J_{\omega_i}\vspace{+1mm}\\
			\mathbf{t}_{y_{i}}^T {J_{v_i}} - r \mathbf{t}_{x_{i}}^T J_{\omega_i}
		\end{bmatrix}, \! & A_{i3} = \begin{bmatrix} -\mathbf{t}_{x_{i}}^T & \mathbf{t}_{x_{i}}^T \hat{p}_{oc_i}\vspace{+1mm}\\
			-\mathbf{t}_{y_{i}}^T & \mathbf{t}_{y_{i}}^T \hat{p}_{oc_i}\end{bmatrix}
	\end{align*}
	with $J_{v_i}=J_{v_i}(\mathbf{q}_i)\in\mathbb{R}^{3 \times n_i}$, $J_{\omega_i}=J_{\omega_i}(\mathbf{q}_i)\in\mathbb{R}^{3 \times n_i}$ being the translational and rotational Jacobian matrices, \begin{math} \mathbf{p}_{oc_i} = \mathbf{p}_{c_i} - \mathbf{p}_{o}\end{math} and $\hat{p}_{{oc}_i}$ is the skew-symmetric matrix formed by the elements of $\mathbf{p}_{{oc}_i}$. The symbol ``$\ \ \widehat{}\ \ $'' over a vector denotes the above diffeomorphism.
		
	The first equation in (\ref{cons}) is the contact constraint implying that the fingertip cannot leave the object's surface. This is true as the system is considered after initial contact with the object is established and as long as the proposed controller applies contact forces inwards the object. The second equation in (\ref{cons}) is the rolling constraint denoting that the velocity of the contact point on the fingertip surface is equal to the velocity of the contact point on the object surface; i.e. that the friction at the contact is sufficient to sustain the tangential contact forces for the rolling motion. 
	
	The system dynamics under the contact and rolling constraints (\ref{cons}) is described by the following equations for both fingers and the object:
	\begin{align}
		M_i(\mathbf{q}_i)\mathbf{\ddot{q}}_i &+ C_i(\mathbf{q}_i,\mathbf{\dot{q}}_i) \mathbf{\dot{q}}_i + {D_{ii}}^T f_i + {A_{ii}}^T \boldsymbol{\lambda}_i, \nonumber \\
		&+ J_{\omega_i}^T K_{s_i} Q_s \boldsymbol{\omega}_{rel_i} = \mathbf{u}_i \label{sys1}
		\end{align}
		\begin{align}
		M \begin{bmatrix} \mathbf{\ddot{p}}_o \\ \boldsymbol{\dot{\omega}}_o \end{bmatrix} & + {D_{13}}^T f_1 + {D_{23}}^T f_2 + {A_{13}}^T \boldsymbol{\lambda}_1 + {A_{23}}^T \boldsymbol{\lambda}_2 \nonumber \\
		& - \begin{bmatrix} 0_{3 \times 3} \\ K_{s_1} \end{bmatrix} Q_s \boldsymbol{\omega}_{rel_1} - \begin{bmatrix} 0_{3 \times 3} \\ K_{s_2} \end{bmatrix} Q_s \boldsymbol{\omega}_{rel_2} = 0, \label{sys2}
	\end{align}
	where $M_i(\mathbf{q}_i)\!\!\in\!\!\mathbb{R}^{n_i \times n_i} \!$,  $C_i(\mathbf{q}_i,\mathbf{\dot{q}}_i) \mathbf{\dot{q}}_i\in\mathbb{R}^{n_i}$ denote the positive definite inertia matrix and the vector of Coriolis and centripetal forces of the $i^{\rm th}$ finger respectively. The lagrange multipliers $f_i$ and $\boldsymbol{\lambda}_i = \begin{bmatrix} \lambda_{y_i} & \lambda_{z_i} \end{bmatrix}^T$are associated with the contact and rolling constraints, respectively. $\mathbf{u}_i\in\mathbb{R}^{n_i}$ is the vector of applied joint torques, $M\!\!=\!\!{\rm diag}\left( m_o I_3,I_o \right)$ is the object's positive definite inertia matrix with object mass $m_o$ and moment of inertia matrix $I_o$, with $I_3$ the identity matrix of size 3. The relative rotational velocity between a fingertip and the object is $\boldsymbol{\omega}_{rel_i} = \boldsymbol{\omega}_{t_i} - \boldsymbol{\omega}_{o} \in\mathbb{R}^{3}$, with rotational velocities \begin{math} \boldsymbol{\omega}_{t_i} \in\mathbb{R}^{3}\end{math}. The friction coefficients $K_{s_i}$ are a diagonal matrix associated with the object's spinning motion which may occur around the interaction line $\dfrac{\mathbf{p}_{c_1}-\mathbf{p}_{c_2}}{\| \mathbf{p}_{c_1}-\mathbf{p}_{c_2} \|} \triangleq \overrightarrow{\mathbf{c}_1\mathbf{c}_2}$ and $Q_s$ is the projection matrix on the interaction line which $Q_s = \overrightarrow{\mathbf{c}_1\mathbf{c}_2} \overrightarrow{\mathbf{c}_1\mathbf{c}_2}^T$.
	
	\subsection{Controller for stable pinching} \label{pinch}
	
\begin{figure}[h!]
	\vspace{-.5em}
		\centering
		\subfloat{\includegraphics[width=0.35\textwidth,trim={100 90 100 100},clip]{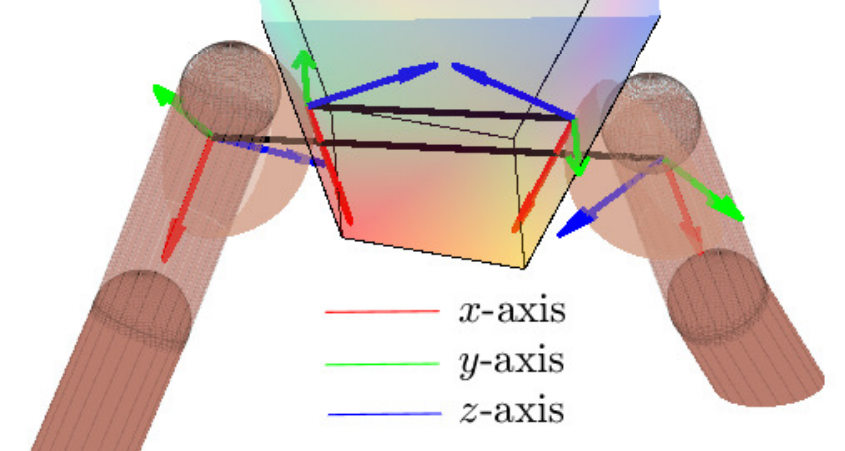}}
		\\			
		\vspace{-.5em}
		\subfloat{\includegraphics[width=0.35\textwidth,trim={70 130 40 190},clip]{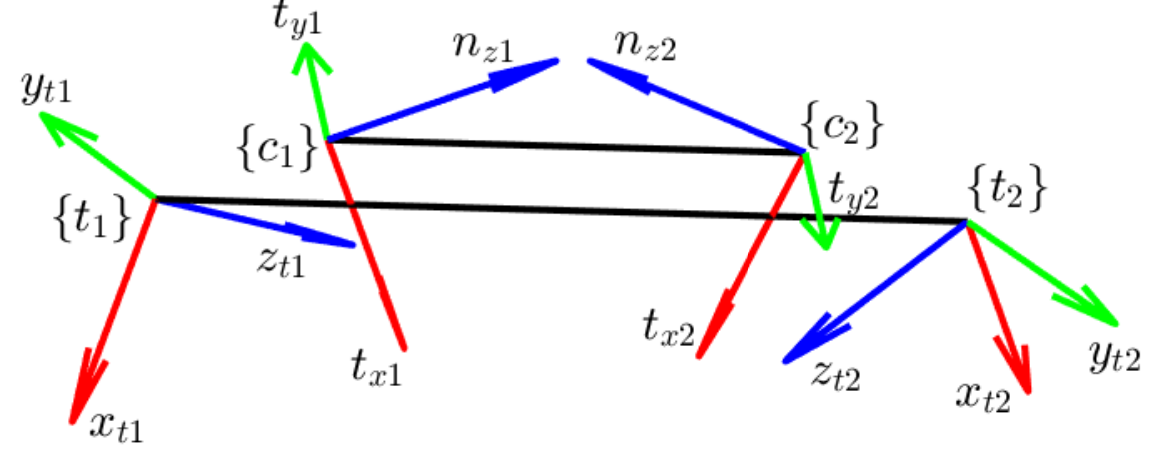}}
		\vspace{-5pt}
		\caption{Contact point $\{c_i\}$ and fingertip $\{t_i\}$ frames at the equilibrium state of grasping a trapezoidal object. The lines connecting the contact points and the fingertips are parallel.}
		\label{f2}
	\end{figure}

	The rolling motion of the fingertips on the contact surface can be separated into two tangential components. The two tangential directions $\mathbf{t}_{x_{i}}$, $\mathbf{t}_{y_{i}}$ and their corresponding contact normals $\mathbf{n}_{z_i}$ define the contact frames (see Fig. \ref{f2}).
	
	The rolling motion of each fingertip on the contact surface can be analysed into its two components as follows: 
	\abovedisplayskip=1pt
	\belowdisplayskip=1pt
	\begin{itemize}
		\item Rolling in the $\mathbf{t}_{x_{i}}$ direction, around $\mathbf{t}_{y_{i}}$ by angle $\phi_i$.
		\item Rolling in the $\mathbf{t}_{y_{i}}$ direction, around $\mathbf{t}_{x_{i}}$ by angle $\psi_i$.
	\end{itemize} 
	These angles $\phi_i$, $\psi_i$ represent the rolling distances covered by each fingertip $i=1,2$: $\dot{\phi}_1 = \mathbf{t}_{y_{1}}^T \boldsymbol{\omega}_{t_1}$, $\dot{\phi}_2 = - \mathbf{t}_{y_{2}}^T \boldsymbol{\omega}_{t_2}$, $\dot{\psi}_1 = \mathbf{t}_{x_{1}}^T \boldsymbol{\omega}_{t_1}$, $\dot{\psi}_2 = \mathbf{t}_{x_{2}}^T \boldsymbol{\omega}_{t_2}$.
	
	We propose the following grasping controller for each finger's joint torques:
	\abovedisplayskip=1pt
	\belowdisplayskip=1pt
	\begin{align}
		\nonumber \mathbf{u}_i = &-K_{vi} \mathbf{\dot{q}}_i - (-1)^i f_d J_{vi}^T \frac{\mathbf{p}_{t_2}-\mathbf{p}_{t_1}}{\| \mathbf{p}_{t_2}-\mathbf{p}_{t_1} \|} \\
		&- r f_d J_{\omega i}^T \big[ (-1)^{i+1} \mathbf{t}_{x_{i}} \sin \psi - \mathbf{t}_{y_{i}} \sin \phi \big], \label{control}
	\end{align}
	where
	\begin{align}
		\label{psiall} &\psi = \psi_1 - \psi_2,	& \phi = \phi_2 - \phi_1  
	\end{align}
	and \begin{math} \mathbf{p}_{t_i}\end{math} is the fingertip position vector.
	
	The controller's tunable parameters are $f_d$ which is a positive constant setting the desired grasping force magnitude at each contact and $K_{vi}\in\mathbb{R}^{n_i \times n_i}$ which is a positive definite diagonal matrix denoting the damping gain of each joint and its values are chosen empirically such that, given a desired $f_d$, the system's performance is as smooth as possible.
	
	The first term of (\ref{control}) is introduced for joint damping. The second term represents applied forces of magnitude $f_d$ at the direction of the line connecting the fingertips $\overrightarrow{\mathbf{t}_1\mathbf{t}_2} \triangleq \dfrac{ \mathbf{ p}_{t_2} -\mathbf{p}_{t_1} } {\| \mathbf{ p}_{t_2} -\mathbf{p}_{t_1} \|}$ and the third term expresses the tangential contact forces at equilibrium.
	
    The closed-loop system under the proposed controller as proved in the Appendix is passive and asymptotically converges to an equilibrium state manifold that satisfies the following conditions:\\
	\noindent 1) The line connecting the centre of the fingertips $\overrightarrow{\mathbf{t}_1\mathbf{t}_2}$ is parallel to the line connecting the contact points $\overrightarrow{\mathbf{c}_1\mathbf{c}_2}$.\\
	\noindent 2) The contact forces that are applied along the $\overrightarrow{\mathbf{t}_1\mathbf{t}_2}$ direction have a magnitude of $f_d$ $N$.\\
	\noindent 3) At equilibrium $\psi = \beta_{\psi}$ and $\phi = \beta_{\phi}$ with $\sin \beta_{\psi} = \mathbf{t}_{y_{i}}^T \overrightarrow{\mathbf{t}_1\mathbf{t}_2}$, $\sin \beta_{\phi} = (-1)^{i+1} \mathbf{t}_{x_{i}}^T \overrightarrow{\mathbf{t}_1\mathbf{t}_2}$.
	
	Notice that the equilibrium state manifold describes antipodal grasps along $\overrightarrow{\mathbf{c}_1\mathbf{c}_2}$ of $f_d$ magnitude. Hence, the proposed controller achieves convergence to an equilibrium state in this manifold without presetting or measuring positions of contacts $\{c_1\}$, $\{c_2\}$. We remark that as there are multiple equilibrium states that satisfy the above conditions, different initial states lead to different equilibrium states in the manifold. In consequence, the initial state of the system does not necessarily correspond to an equilibrium. The closed-loop system's passivity means that after any non-persistent disturbance the system will always converge to a state in the equilibrium manifold.
	
	Further notice that the proposed control law in equations~(\ref{control},\ref{psiall}) assumes knowledge of the tangential directions at the contacts; therefore an estimate of the contact surface orientation is needed, which is here achieved by using tactile sensing. This is in contrast to our previous work where tactile sensing was not utilised \cite{psomopoulou_karashima_doulgeri_tahara_2018}. However, we will show that the controller is robust to measurement errors of the contact surface orientation. Other quantities are calculated using only the robotic finger forward kinematics and the radius of the hemispherical tips.
	
	The proposed control law dynamically stabilises the grasp of an object of any shape via fingertip rolling on the contact surfaces with a desired grasp force $f_d$. The $f_d$ parameter should be set so that assumption (ii) holds and contact forces stay within the friction cone. The controller does not assume solving the inverse kinematics of the system for trajectory planning neither does it depend on the knowledge of its dynamic parameters. 
	
	\subsection{Tactile data acquisition and processing for the Shadow Modular Grasper} \label{dnn}
	
	The Shadow Modular Grasper is a fully-actuated three-fingered robotic hand with 9 DoF (3 DoF per finger). The hand has a payload of 2\,kg, with each finger applying up to 10\,N of normal continuous force and each joint has a dedicated torque sensor for a 10\,kHz closed-loop control \cite{Shadow}. The hand is fully integrated with ROS and is provided with open APIs for grasping control. 
	
	Tactile sensing is enabled on the hand by replacing the fingertips of the Modular Grasper with custom-built tactile sensors \cite{8656557} (see Fig \ref{fig1}). The tactile fingertips are adapted from an optical biomimetic tactile sensor developed in Bristol Robotics Laboratory, the BRL TacTip \cite{doi:10.1089/soro.2017.0052}. Deformation of the tactile sensing pad is imaged with the internal camera at its native resolution of 1920 x 1080, then adaptively thresholded with a Gaussian filter (width 27, mean 0 pixels) and subsampled/cropped to 240x135-pixel greyscale images (Fig. \ref{fig1}). All image acquisition and processing was carried out in Python OpenCV. 
	
	\begin{table}[h!]
		\centering
		\begin{tabular}{lc}
			\toprule
			Hyperparameters & Optimised values\\
			\midrule
			$\#$ convolutional hidden layers, $N_{\rm conv}$ & 5 \\
			$\#$ convolutional kernels, $N_{\rm filters}$ & 256  \\
			$\#$ dense hidden layers, $N_{\rm dense}$ & 2 \\
			$\#$ dense hidden layer units, $N_{\rm unit}$ & 64 \\
			hidden layer activation function & eLU \\
			dropout coefficient & 0.01 \\
			L1-regularisation coefficient & 0.0001 \\
			L2-regularisation coefficient & 0.01 \\
			batch size & 16 \\
			\bottomrule
		\end{tabular}
		\caption{Neural network and learning hyperparameters.}
		\label{tbl:deepcnn}
		\vspace{-1em}
	\end{table}
	
	\begin{figure}[h!]
		\centering
		\begin{tabular}{@{}c@{}}
			\includegraphics[width=0.8\columnwidth,trim={0 0 250 300},clip]{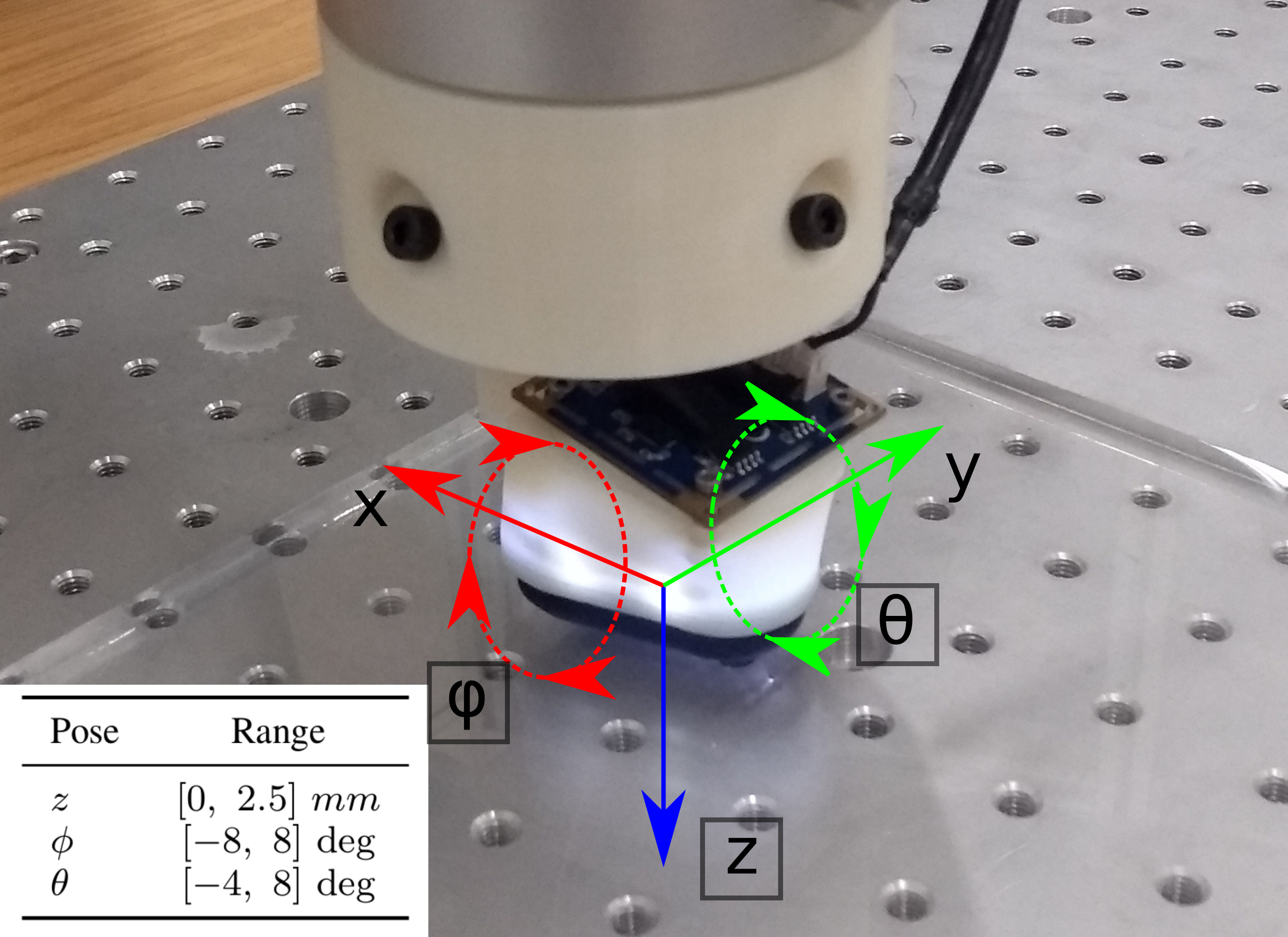}
		\end{tabular}
		\vspace{-5pt}
		\caption{Data collection set-up with tactile fingertip mounted as an end-effector on a ABB robot arm. The pose components collected are roll $\phi$, pitch $\theta$ and depth $z$ relative to the sensor frame. Pose ranges used as labels to train/test the neural network are shown in the table.}
		\vspace{0em}
		\label{fig2}	
	\end{figure}
	
	The tactile image data was used to estimate the pose from a contacted surface relative to the sensor (Fig. \ref{fig2}) and thus estimate the contact frame orientation of each finger with respect to the hand base. The tactile image data was also used for contact deformation measurement. A simple yet robust measure of the difference in tactile images can be found from the Structural Similarity Index Measure (SSIM) \cite{8460494}, which can be used to measure contact deformation by comparing a tactile image against a non-deformed reference image \cite{james2020tactile}. Here we use $e_{\rm SSIM}(I) = 1 - {\rm SSIM}(I, I_{\rm ref})$ as a measure of the deformation of image $I$ compared with the reference image $I_{\rm ref}$, with SSIM implemented using Python SciKit-Image and computed from the local means, variances and cross-covariance of the two images \cite{1284395}. The SSIM-based deformation measure changed gradually as the contact intensified, making it suitable for use in a feedback controller.
	
	The fingertip is mounted as an end-effector on a 6 DoF robot arm (IRB120, ABB Robotics). The fingertip then repeatedly contacted a flat surface to gather labelled contact data to train, validate and test a deep neural network. For details of the deep learning method used, we refer to \cite{9058673}. Each of the 5000 samples of data had a random labelled pose (ranges in Table inside Fig. \ref{fig2}). The network hyperparameters were optimised (Table \ref{tbl:deepcnn}) with the training implemented in the Python Keras library using a GeForce GTX 1660 GPU. This work considers a two-fingered grasp; thus, the above model procedure was repeated for both tactile fingertips.
	\vspace{-5pt}
	
	\begin{figure}[h!]
	\centering
	\begin{tabular}{@{}c@{}}
		\includegraphics[scale=0.25,trim={0 0 0 0},clip]{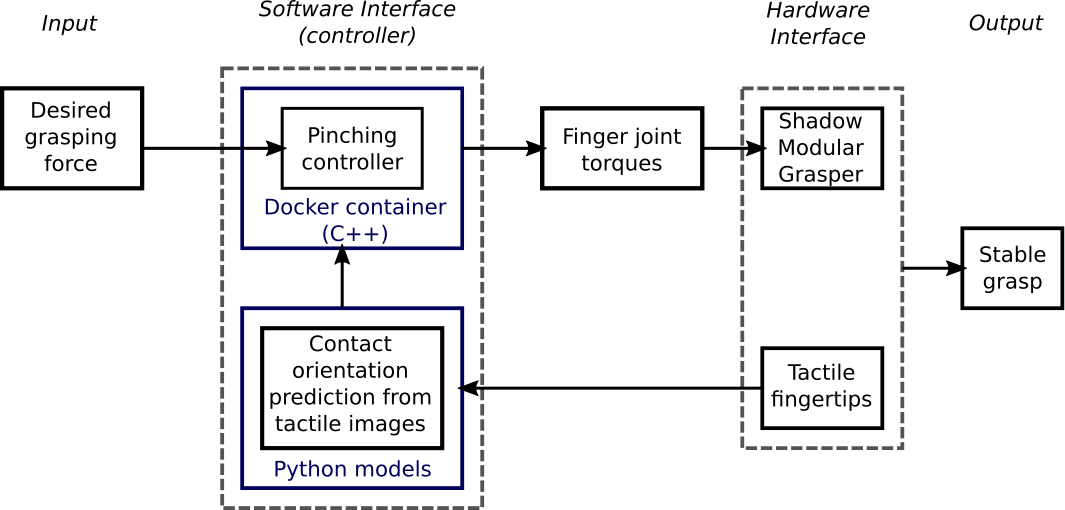}
	\end{tabular}
	\caption{Block diagram of the proposed controller with feedback derived from the tactile images.}
	\vspace{-1em}
	\label{fig3}	
    \end{figure}
    
	\subsection{System Integration}

	For the purpose of this study, we use the predicted contact orientation from the tactile sensor (Section \ref{dnn}) inside the controller of Section \ref{pinch}. To that end, the predicted Euler angles are transformed into the unit vectors $\mathbf{t}_{x_{i}}$ and $\mathbf{t}_{y_{i}}$ and expressed in the Shadow Modular Grasper's base frame. The predicted sensor deformation $z$ is used to detect contact with an object. Two Python models, one per sensor, run on the host PC and interact with the grasp controller (in C++) which runs in a Docker container via a ROS-network (Fig. \ref{fig3}).
	
	Each finger's joint can be controlled in either position or torque mode. The control is implemented within a 1\,kHz update loop running inside the Docker container. The two fingertip poses are predicted with an update rate of $\sim$100\,ms and then broadcasted to the grasping controller inside the Docker container via ROS topics.
	
	\begin{figure*}[t!]
		\centering
		\begin{tabular}[b]{c}
			{\bf (a) Initial Configuration } \\
			\hspace{0cm}\includegraphics[width=0.77\textwidth,trim={20 85 20 25},clip]{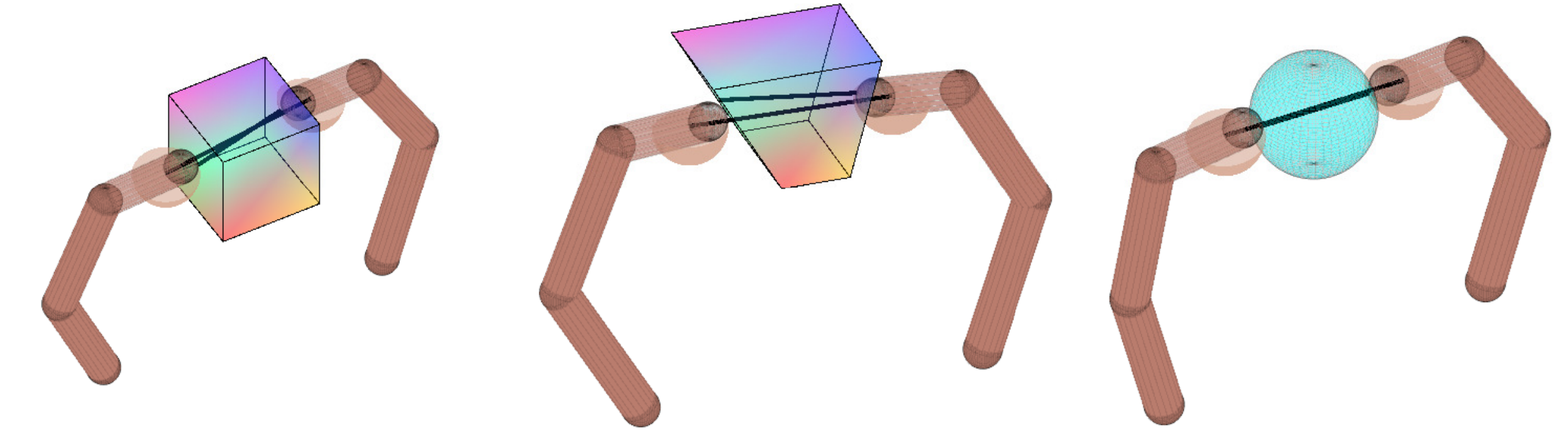} \\
			{\bf (b) Final Configuration } \\
			\hspace{0cm}\includegraphics[width=0.77\textwidth,trim={10 50 70 25},clip]{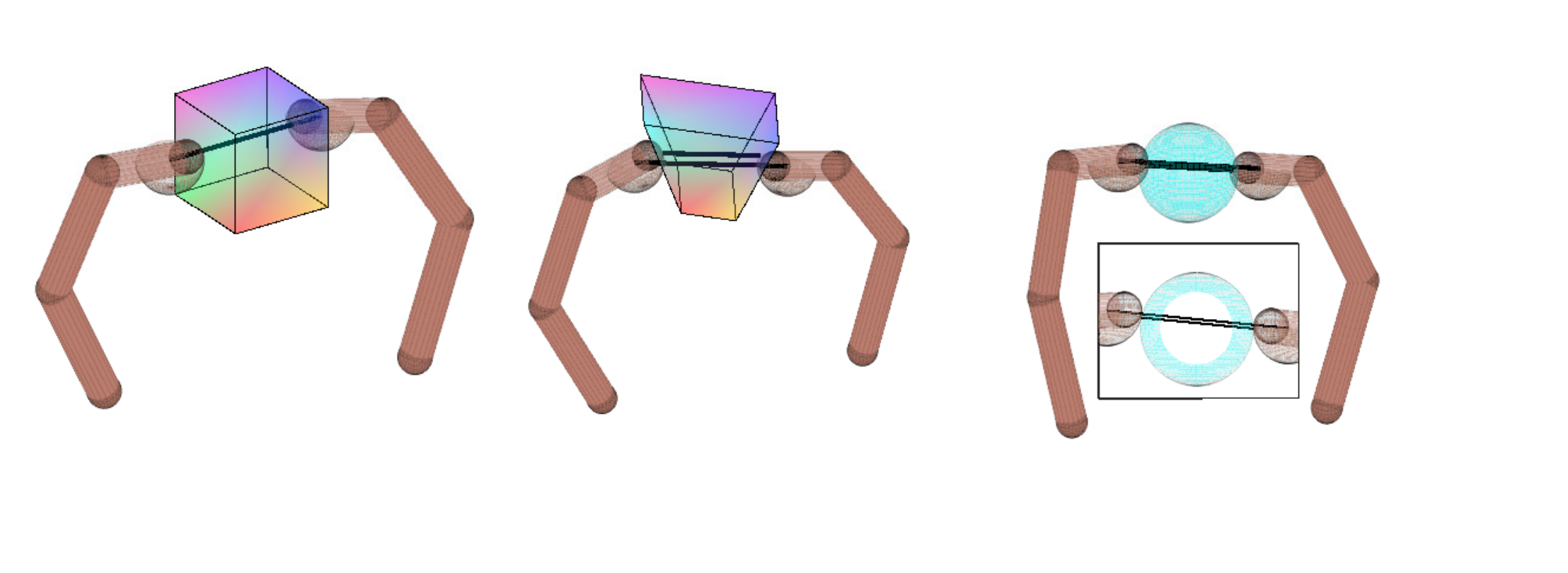} \\
		\end{tabular}
		\vspace{-5pt}
		\caption{Initial configuration and system equilibrium state for the pinching controller in simulation for three different objects of mass $0.0021$ kg. A cube of side length $0.048$ m, a trapezoidal object (height $0.048$ m, small base $0.0277$ m side angles $30$ \& $15$ $\deg$) and a sphere of radius $0.024$ m (zoom-in detail of the parallel lines)}%
		\vspace{-1em}
		\label{sys00}%
	\end{figure*}

    A grasp is comprised of two distinct phases:\\
    \noindent {\em (i) Closing phase} During this phase all joints are controlled in position mode and commanded to a set of target angles via a PID which is an implementation of the rosControl ROS-package. The hand controller switches to the next phase once contact is detected using the SSIM (Section \ref{dnn}).\\
    \noindent {\em (ii) Grasping phase} After both sensors have detected contact, all joints switch to torque mode and the grasping controller (Section \ref{pinch}) is activated, which leads to the hand stably grasping the object.

\begin{figure}[b!]
		\centering
		\includegraphics[scale=.5,trim={9 17 0 23},clip]{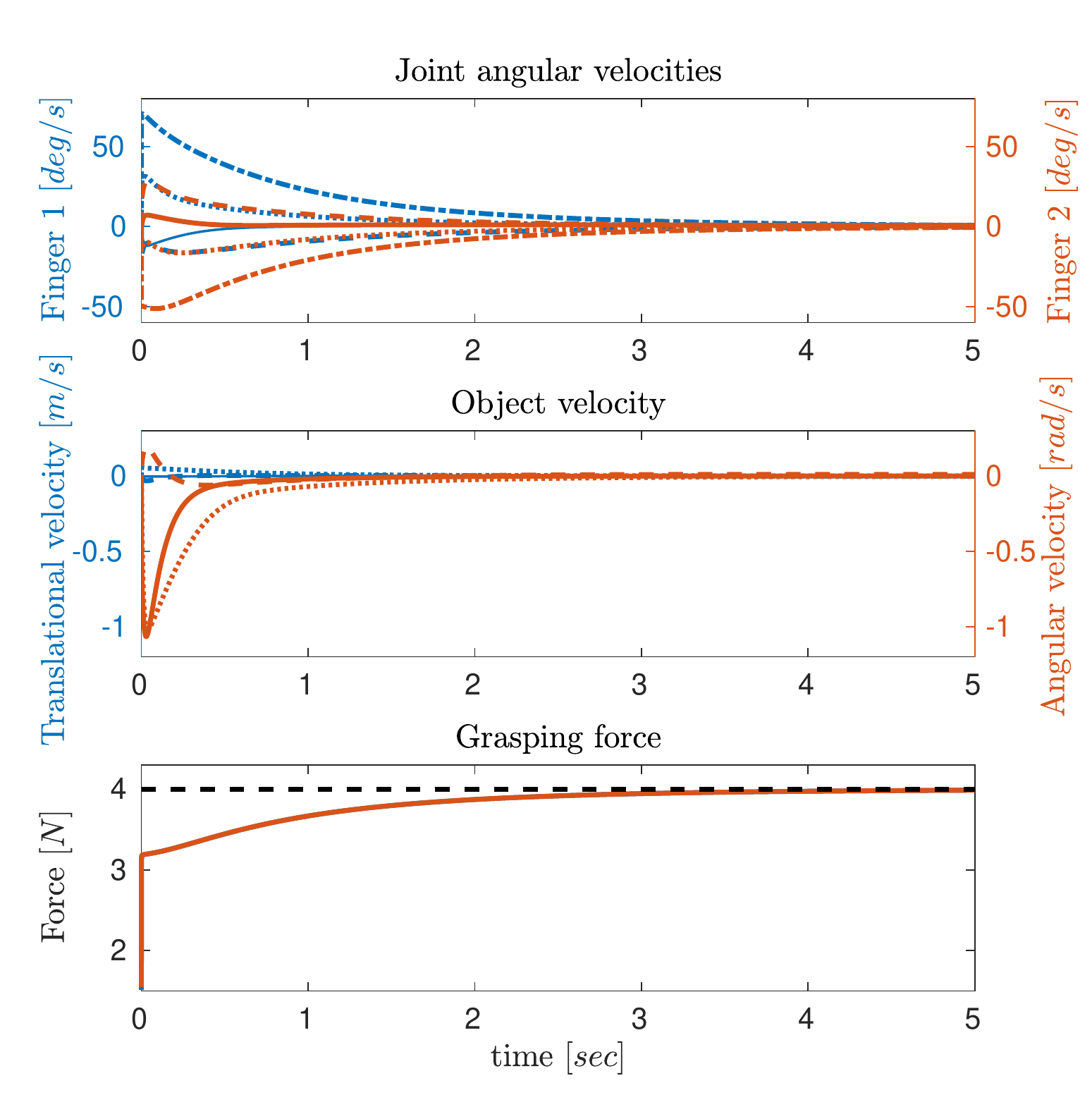}
		\caption{Finger joint and object velocities for the trapezoidal object as well as grasping force response.}
		\label{qdot}
\end{figure}

\begin{figure}[b!]
		\vspace{-1em}
		\centering
	    \includegraphics[width=\columnwidth,trim={0 65 0 0},clip]{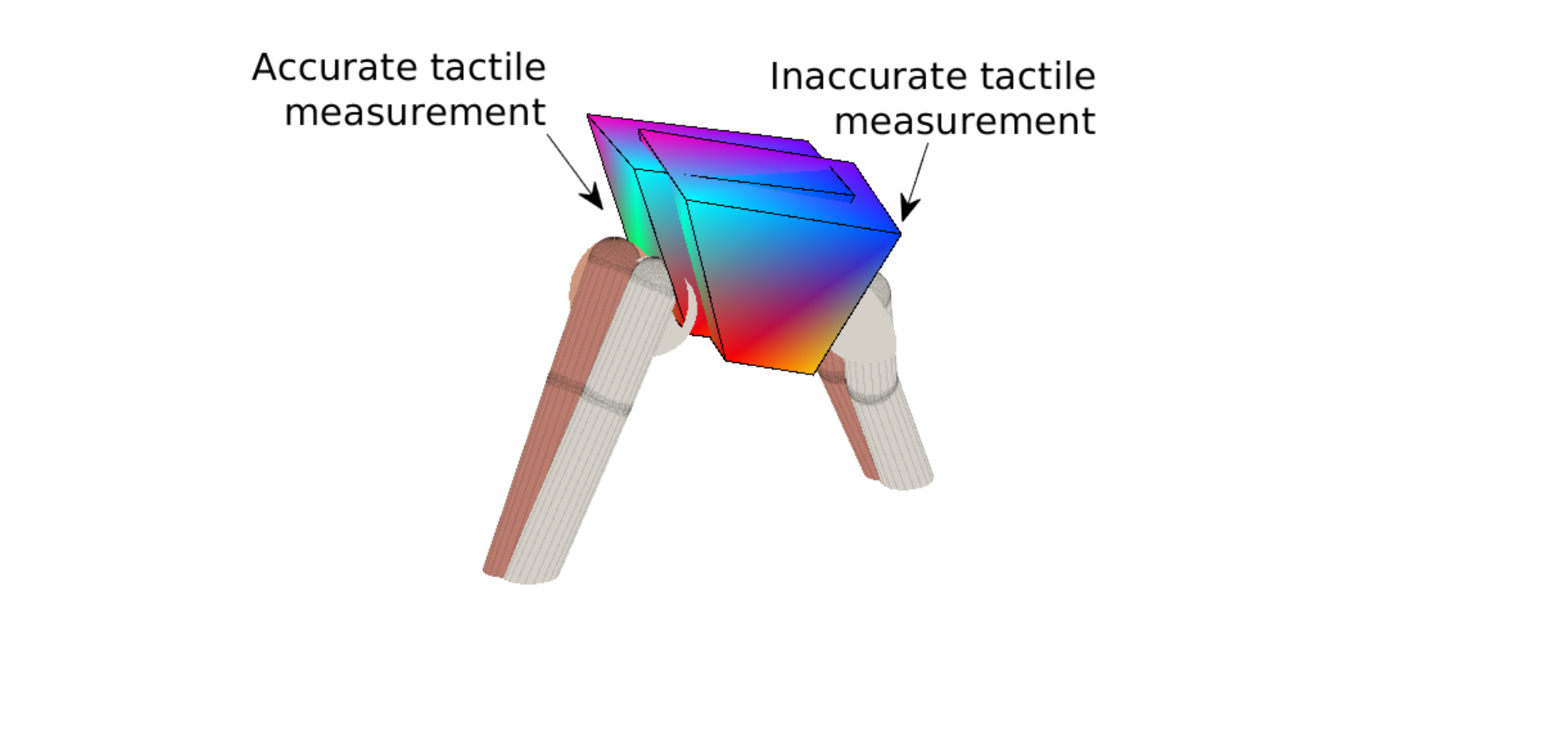} \\
		\caption{Equilibrium configuration without (brown fingers) and with (grey fingers) tactile measurement errors  of $30^\circ,15^\circ$ per contact surface orientation.}%
		\label{sysfp090err}%
	\end{figure}

	\section{RESULTS}

	\subsection{Simulation of stable pinching} \label{sim}
	
	Two identical robotic fingers with rigid fingertips (radii $r=0.015$\,m) are simulated in MATLAB. Three different object shapes are used; a cube, a trapezoidal object and a sphere. They are considered rigid and their parameters are given in the caption of Fig. \ref{sys00}. The system is simulated under the proposed controller with damping gains $K_{v_i} = 0.07I_4$ for both fingers $i = 1,2$ and desired grasping force $f_d=4\,$N.
	
	The initial system configuration is shown in the top row of Fig. \ref{sys00}. The bottom row shows the system's equilibrium state where it is clear that the lines connecting the fingertips and the contact points are parallel. System time responses are shown indicatively for the trapezoidal object in Fig.~\ref{qdot} and are consistent with theoretical expectations. Joint and object velocities converge to zero and the grasping force converges to the desired magnitude $f_d = 4$\,N. 
	
	\begin{figure*}[t]
		\centering
		\begin{tabular}{@{}c@{}}
			\vspace{.2em}\hspace{-.4em}			
			\includegraphics[width=0.5\columnwidth,trim={0 0 0 0},clip]{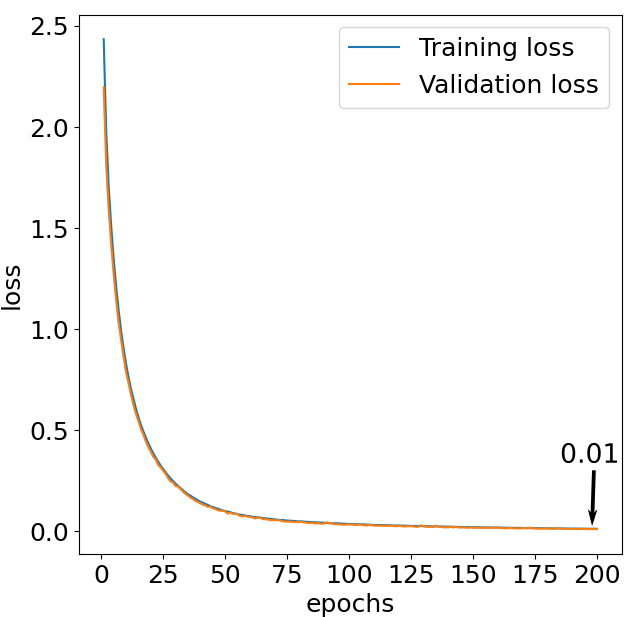}
			\includegraphics[width=1.4\columnwidth,trim={0 0 0 0},clip]{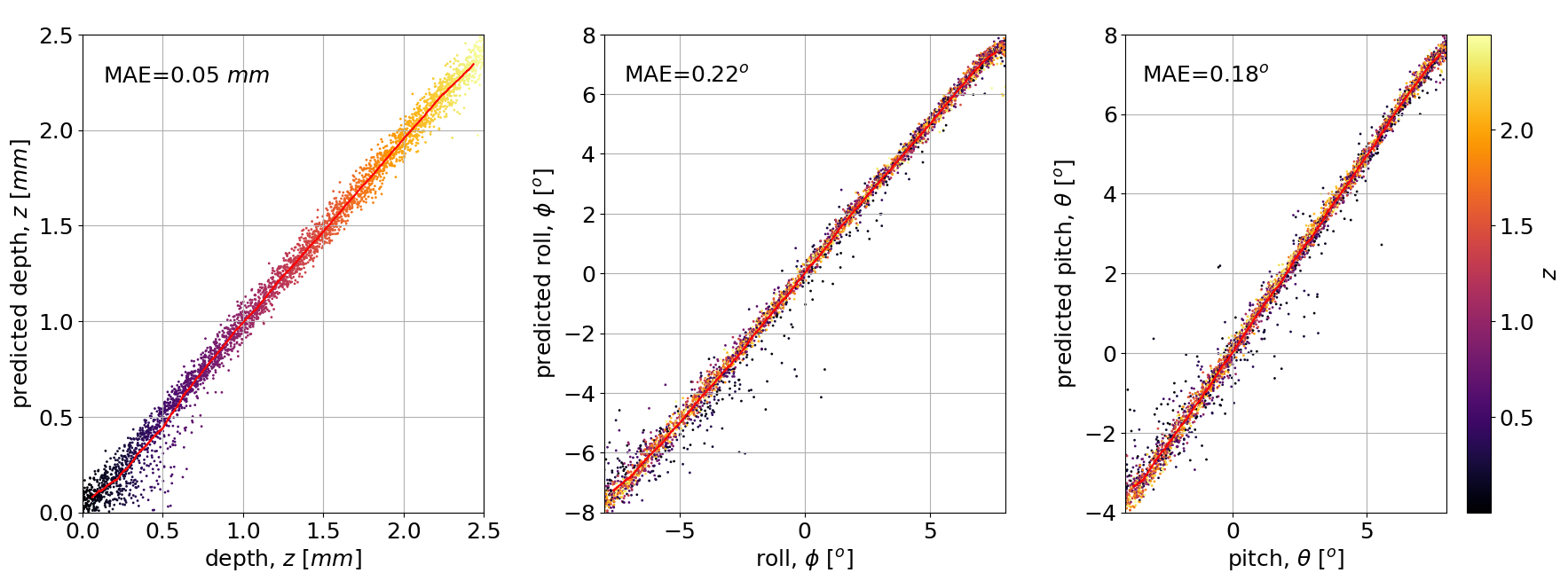} \\	
		\end{tabular}
		\caption{Tactile sensor's training and validation loss as well as performance of the trained models for depth, roll and pitch.}
		\label{fig5}
		\vspace{-0.5em}
	\end{figure*}
	
	\begin{figure*}[t!]
		\centering
		\begin{tabular}[b]{c}
			{\bf (a) Initial Configuration } \\
			\hspace{0cm}\includegraphics[width=0.9\textwidth,trim={0 0 0 0},clip]{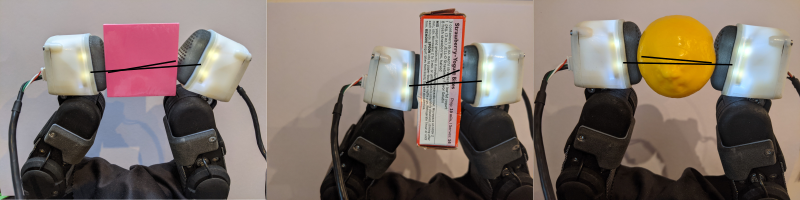} \\
			{\bf (b) Final Configuration } \\
			\hspace{0cm}\includegraphics[width=0.9\textwidth,trim={0 0 0 0},clip]{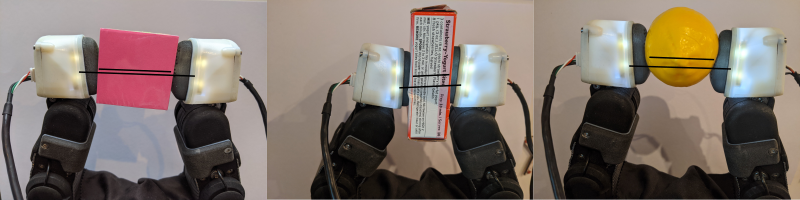} \\
		\end{tabular}
		\caption{Initial configuration and system equilibrium state for the pinching controller on the tactile Shadow Modular Grasper for three different objects, a stack of post-it notes, an empty cardboard box and a plastic lemon.}%
		\label{real}%
		\vspace{-1em}
	\end{figure*}
	
	The only external measurements used by the proposed controller come from the tactile sensing. Therefore, we investigate the robustness to measurement errors in the contact surface orientation. To demonstrate this, we simulate a measurement error of $30^\circ,15^\circ$ for each contact surface orientation respectively. This is implemented for the trapezoidal object, but practically it is like the fingers `feel' they are holding a cube. Simulation results show that the system converges to a different equilibrium state of the manifold with respect to the final finger-object pose (Fig. \ref{sysfp090err}, equilibrium configurations with (grey fingers) and without (brown fingers) errors are overlayed).
	
	\subsection{Tactile perception of contact surface orientation}
	
	\begin{table}[b!]
		\centering
		\begin{tabular}{cccc}
			\toprule
			Pose component & Sensor 1 MAE & Sensor 2 MAE & range\\
			\midrule
			vertical $z$ & $0.05$\,mm & $0.08$\,mm & $2.5$\,mm\\
			roll $\phi$ & $0.22^\circ$ & $0.45^\circ$ & $16^\circ$  \\
			pitch $\theta$  & $0.18^\circ$ & $0.38^\circ$ & $12^\circ$\\
			\bottomrule
		\end{tabular}
		\caption{Pose prediction performance.}
		\label{tbl:performance}
	\end{table}
	
	Two deep convolutional neural networks were trained to regress pose over the labelled tactile images (Section \ref{dnn}). The training involved Bayesian optimisation over the network architecture and hyperparameters \cite{9058673} (optima reported in Table \ref{tbl:deepcnn}). Pose estimation was then assessed on distinct test sets for each sensor. Plots of the predictions versus ground truth show good performance for all pose components ($z$, $\phi$, $\theta$) (Fig. \ref{fig5}). As is evident from errors due to small vertical pose component (Fig. \ref{fig5} colourmap according to $z$), the outliers are caused by little or no contact with the surface. In the scope of this paper, information from the sensors is only needed when there is contact with the object. The overall accuracy is similar for both sensors and typically below 0.1\,mm and 0.5$^\circ$ (Table \ref{tbl:performance}). Small differences between the two sensors are negligible with respect to their ranges and can be attributed to variations in their fabrication and optics, such as the amount of injected gel and the camera placement. These variations are also a reason for training a network for each sensor.
	
	\subsection{Stable pinching experimental results on the tactile Shadow Modular Grasper}
	
\begin{figure}[h!]
		\centering
		\includegraphics[scale=.75,trim={0 0 0 0},clip]{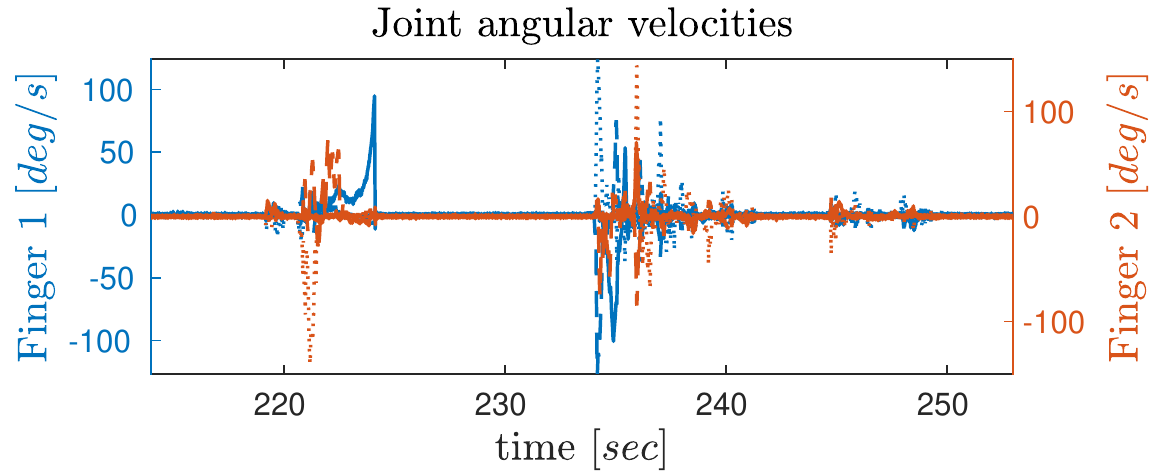}
		\caption{Joint angle velocities with external perturbation while grasping an object. Initial grasp at $220$ sec, large perturbation at $235$ sec, smaller perturbation at $245$ sec.}
		\vspace{-1em}
		\label{perturbation}
	\end{figure}

	We consider four different objects of different shapes and softness (Figs \ref{fig1}, \ref{real}-\ref{perturbation}). The system runs under the proposed controller with damping gains $K_{v_i} = 0.07I_4$, desired grasping force $f_d=10$\,N and fingertip radius $r = 0.015$\,m.
	
	The controller achieves a stable grasp by rolling the fingertips on the contact surfaces. The finger configuration is kept as close as possible to the initial one. Figure \ref{real} shows the initial configuration and the system's equilibrium state for households objects whose shapes resemble those used in simulation (Section \ref{sim}). Note the empty cardboard box in the middle deforms under grasping into a trapezoid. The experiments in this section show that the controller achieves stability even in the case of soft materials. 
	
	Figure \ref{fig1} shows the system's equilibrium state for the brain-shaped object which is the most challenging to control because it is soft and has an uneven texture. The fingertip and contact frames are captured from the online rendition of the frames in RVIZ, a 3D visualisation tool of ROS, and are overlaid on the experimental setup photo. 
	
	In both Figs \ref{fig1} and \ref{real}, it is clear that the lines connecting the fingertips and contact points are parallel, which matches the theoretical results from Section \ref{pinch}. To further test the stability of the system, two external perturbations were applied by pushing the right finger just before $t = 235$\,sec and $t = 245$\,sec. Fig. \ref{perturbation} shows the finger joint angle velocities which return to zero after perturbations; hence, the system converges to an equilibrium state. 
	
	\section{DISCUSSION}
	
	In this paper, we have presented a dynamically stable controller for pinching an arbitrary-shaped object with two robotic fingers. The novelty with respect to our previous work in this area~\cite{psomopoulou_karashima_doulgeri_tahara_2018} lies in the use of tactile sensing to address grasp stability in a 3D setting. This was attained by using a convolutional neural network to extract the orientation of the contact frames for use in the controller. We showed that the system is robust to tactile measurement errors, in that a systematic error in estimating the contact orientation leads to a new equilibrium nearby the original system pose. While our tactile measurements were accurate on test data (to within 0.5$^\circ$), systematic measurement errors could occur on objects that are distinct from the planar surface used in training, such as for very curved or soft regions of the object. That objects of these types (a squidgy brain and lemon) were stably grasped indicates the method's robustness.  
	The control also appeared to be robust to other sources of error. In particular, the theoretical analysis of the system was based on the assumptions that the fingertips and objects are rigid and that the mass of the object is small enough to ignore gravity. However, the controller performed well with soft, curved tactile fingertips and the objects ranged in softness, curvature and weight.
	
	The controller's performance depends however on the assumption of rolling fingertips. In practice, this means that the friction at each contact is such that the contact force remains in the friction cone during the grasp so that there is no slippage in the fingertip-object contact and hence rolling is possible. The value of $f_d$ as well as soft contacts by the deformable fingertips play a role in satisfying this constraint. If a non-satisfaction of assumptions causes slippage (eg. heavy object, persistent external perturbation, $\overrightarrow{\mathbf{c}_1\mathbf{c}_2}$ outside of the friction cones at the contacts), the grasp will fail.
	
	Finally, although the scope of the present work was stable pinch grasping, we would like to emphasise that the controller should extend to a broader range of capabilities, including stable in-hand manipulation of unknown objects using touch. By introducing additional control parameters in the rolling angles $\phi$ and $\psi$ of (\ref{psiall}), the rolling of the fingertips on the contact surface can be controlled to manipulate the object's in-hand orientation. In consequence, the controller could be used to adjust the hand-object composite to the environment's geometry, for example to place an object inside a constrained space. It could also be used to adjust the internal force to ensure the safety of the grasp by increasing the grasping force in case of slippage~\cite{9252140}. These topics will be examined in future work, along with extending the controller to use multiple fingers of the robot hand. 
    \vspace{-5pt}
	
	\section*{Appendix: System Equilibrium and Stability Analysis} \label{stability}

	Substituting (\ref{control}) into (\ref{sys1}), the closed loop system can be written as:
	\begin{align}
		\nonumber M_i \mathbf{\ddot{q}}_i &+ (C_i + K_{vi})\mathbf{\dot{q}}_i + J_{\omega_i}^T K_{s_i} Q_s \boldsymbol{\omega}_{rel_i} + D_{ii}^T \Delta f_i \\
		\label{cl1} & + A_{ii}^T \begin{bmatrix}
			\Delta \lambda_{x_i}\\
			\Delta \lambda_{y_i}
		\end{bmatrix}  + r J_{\omega_i}^T \Delta \mathbf{N}_i = 0\\
		\label{cl2} M_o \mathbf{\ddot{p}}_o &- \sum_{i=1}^{2}(\mathbf{n}_{z_i} \Delta f_i + \mathbf{t}_{x_{i}} \Delta \lambda_{x_i} + \mathbf{t}_{y_{i}} \Delta \lambda_{y_i}) = 0\\
		\nonumber I_o \boldsymbol{\dot{\omega}}_o & + \sum_{i=1}^{2}\hat{p_{oc_i}}^T(\mathbf{n}_{z_i} \Delta f_i + \mathbf{t}_{x_{i}} \Delta \lambda_{x_i} + \mathbf{t}_{y_{i}} \Delta \lambda_{y_i})\\
		\label{cl3} & - \sum_{i=1}^{2} K_{s_i} Q_s \boldsymbol{\omega}_{rel_i} + \mathbf{S}_N = 0
	\end{align}
	where $\Delta f_i = f_i - (-1)^{i+1}f_d \mathbf{n}_{z_i}^T \overrightarrow{\mathbf{t}_1\mathbf{t}_2}$, $\Delta \lambda_{x_i}=\lambda_{x_i} - (-1)^{i+1}f_d \mathbf{t}_{x_{i}}^T \overrightarrow{\mathbf{t}_1\mathbf{t}_2}$, $\Delta \lambda_{y_i}= \lambda_{y_i} - (-1)^{i+1}f_d \mathbf{t}_{y_{i}}^T \overrightarrow{\mathbf{t}_1\mathbf{t}_2}$, $\Delta \mathbf{N}_i=(-1)^{i+1} f_d (\mathbf{t}_{y_{i}} \mathbf{t}_{x_{i}}^T - \mathbf{t}_{x_{i}} \mathbf{t}_{y_{i}}^T) \overrightarrow{\mathbf{t}_1\mathbf{t}_2}+ f_d \big[ (-1)^{i+1} \mathbf{t}_{x_{i}} \sin \psi - \mathbf{t}_{y_{i}} \sin \phi \big]$ and $\mathbf{S}_N=f_d (\hat{p}_{oc_1}^T - \hat{p}_{oc_2}^T)\overrightarrow{\mathbf{t}_1\mathbf{t}_2}$.

The system equilibrium is found by setting velocities and accelerations to zero in (\ref{cl1}-\ref{cl3}). This leads to: $\Delta f_{i\infty} = \Delta \lambda_{y_{i\infty}} = \Delta \lambda_{z_{i\infty}} =\Delta \mathbf{N}_{i\infty}=S_{N\infty} = 0$, which then gives:
\begin{align}
	\big( \hat{p}_{oc_{2\infty}}^T - \hat{p}_{oc_{1\infty}}^T\big) \overrightarrow{\mathbf{t}_1\mathbf{t}_2}_{\infty}  &=  0 \label{kala}\\
	\label{rel1} \psi_{1\infty} - \psi_{2\infty} = \beta_{\psi\infty},&\ \ \  \phi_{2\infty} - \phi_{1\infty} = \beta_{\phi\infty}
\end{align}
with $\sin \beta_{\psi\infty}\!= \mathbf{t}_{y_{i\infty}}^T \overrightarrow{\mathbf{t}_1\mathbf{t}_2}_{\infty}$, $\sin \beta_{\phi\infty}\!= (-1)^{i+1} \mathbf{t}_{x_{i\infty}}^T \overrightarrow{\mathbf{t}_1\mathbf{t}_2}_{\infty}$, and the subscript $_\infty$ denoting equilibrium values.

 Notice that $\mathbf{p}_{oc_2}-\mathbf{p}_{oc_1}\!=\overrightarrow{\mathbf{c}_1\mathbf{c}_2}$ is the interaction line vector; hence (\ref{kala}) indicates a zero outer product of $\overrightarrow{\mathbf{c}_1\mathbf{c}_2}$ with $\overrightarrow{\mathbf{t}_1\mathbf{t}_2}$, which implies that these lines are parallel to each other at equilibrium. Notice also that $\Delta f_{i\infty} = \Delta \lambda_{y_{i\infty}} = \Delta \lambda_{z_{i\infty}}=0$ satisfy the object's translational motion equation (\ref{cl2}) at equilibrium. The physical meaning is that, at equilibrium, since $\overrightarrow{\mathbf{t}_1\mathbf{t}_2} || \overrightarrow{\mathbf{c}_1\mathbf{c}_2}$, contact forces lie on the interaction line with magnitude $f_d$.
	
To facilitate the stability analysis, we rewrite the closed loop system equation in a compact form, collecting all Lagrange multipliers in the vector $\boldsymbol \lambda= [f_1 \  f_2 \  \boldsymbol{\lambda}_1^T \  \boldsymbol{\lambda}_2^T]^T$ and all system position variables in $\mathbf{x} = [\mathbf{q}_1^{T} \ \mathbf{q}_2^{T} \ \mathbf{p}_o^{T} \ \boldsymbol{\omega}_o^{T}]^T$:
	\begin{align}
		\nonumber M_s \mathbf{\ddot{x}}\! &+ \! ( C_s + K_{v} )\mathbf{\dot{x}} \!+\! A \boldsymbol \lambda \!+\! B_{J_\omega} Q \boldsymbol{\omega}_{rel} \!+\! f_d {J_{v}}^T \frac{\mathbf{p}_{t_2} - \mathbf{p}_{t_1}}{\| \mathbf{p}_{t_2} - \mathbf{p}_{t_1} \|} \\
		&+ r f_d {J_{\omega}}^T \begin{bmatrix}
			\mathbf{t}_{x_{1}} \sin \psi - \mathbf{t}_{y_{1}} \sin \phi\\
			- \mathbf{t}_{x_{2}} \sin \psi - \mathbf{t}_{y_{2}} \sin \phi\\
			0_{6 \times 1}
		\end{bmatrix}  = 0  \label{clalt}
	\end{align}
	with $M_s = {\rm diag}\left( M_1, M_2, M \right)$, $C_s = {\rm diag}\left( C_{1}, C_{2}, 0_{6 \times 6} \right)$, $K_v = {\rm diag}\left( K_{v_1}, K_{v_2}, 0_{6 \times 6} \right)$, $J_v = \begin{bmatrix}
			-J_{v1} & J_{v2} & 0_{3 \times 6}
		\end{bmatrix}$, $J_\omega = \begin{bmatrix}
			J_{\omega 1} & J_{\omega 2} & 0_{3 \times 6}
		\end{bmatrix}$, $Q = {\rm diag}\left[ K_{s_1}Q_s,K_{s_2}Q_s \right]$, $\boldsymbol{\omega}_{rel} = \begin{bmatrix} \boldsymbol{\omega}_{rel_1}^T & \boldsymbol{\omega}_{rel_2}^T\end{bmatrix}^T$, $A \!=\!\! \left[ \begin{smallmatrix}
			{D_{11}}^T & 0_{n_1 \times 1}    & {A_{11}}^T & 0_{n_1 \times 2} \\
			0_{n_2 \times 1}   & {D_{22}}^T &   0_{n_2 \times 2}  & {A_{22}}^T \\
			{D_{13}}^T & {D_{23}}^T & {A_{13}}^T & {A_{23}}^T\end{smallmatrix} \right]$, $B_{J_\omega} \!\!=\!\! \left[ \begin{smallmatrix} J_{\omega_1}^T & 0_{5 \times 3} \\
			0_{4 \times 3} & J_{\omega_2}^T \\
			0_{3 \times 3} & 0_{3 \times 3} \\
			-I_3 & -I_3\end{smallmatrix}\right]$.
	Similarly, the constraints can be written as: $A^T \mathbf{\dot{x}} = 0$. Left-multiplying (\ref{clalt}) by $\mathbf{\dot{x}}^{T}$ yields $\dfrac{dV}{dt} + W = 0$, where:\\
	\begin{align*}
		&V = \frac{1}{2} \mathbf{\dot{x}}^{T} M_s \mathbf{\dot{x}} + f_d \|\mathbf{p}_{t_2} - \mathbf{p}_{t_1}\| + r f_d \big(z_1(t) + z_2(t)\big) \\
		&W = \sum_{j=1}^{n_1} k_{v_{1j}} \dot{q}_{1j}^2 + \sum_{j=1}^{n_2} k_{v_{2j}} \dot{q}_{2j}^2 + \sum_{i=1}^{2} \boldsymbol{\omega}_{rel_i}^T K_{s_i} Q_s \boldsymbol{\omega}_{rel_i} 
	\end{align*}
	where  $z_1(t) = \int_0^{\phi} \sin \xi_1  \mathrm{d}\xi_1$, $z_2(t) = \int_0^{\psi} \sin \xi_2  \mathrm{d}\xi_2$ and $k_{v_{ij}}$ are the diagonal elements of the matrix $K_{v_i}$. Clearly $V$ is positive definite with respect to $\mathbf{\dot{x}}$, $\|\mathbf{p}_{t_2} - \mathbf{p}_{t_1}\|$, $z_1(t)$ and $z_2(t)$ for $-\frac{\pi}{2} < \phi < \frac{\pi}{2}$ and $-\frac{\pi}{2} < \psi < \frac{\pi}{2}$ in the constraint manifold defined by $\mathcal{M}_c(\mathbf{x}) = \{\mathbf{x} \in \mathbb{R}^{n_1 + n_2 +6} : A^T \mathbf{\dot{x}} = 0\}$. Given $\dot{V} \!=\! -W \leq 0$, it is clear that $V( t ) \leq V\left( 0 \right)$ holds and consequently $\mathbf{\dot{x}}$, $\|\mathbf{p}_{t_2} - \mathbf{p}_{t_1}\|$, $z_1(t)$ and $z_2(t)$ are bounded. Following the proof line in \cite{psomopoulou_karashima_doulgeri_tahara_2018}, it is proved that the system velocities and accelerations converge to zero $\mathbf{\dot{x}}, \mathbf{\ddot{x}} \rightarrow 0$ and that $\mathbf{\dot{x}}$ converges to zero exponentially as $t \rightarrow \infty$.

	\bibliographystyle{IEEEtran}
	\bibliography{IEEEabrv,Refs}
	
\end{document}